# How LLM Ordering Agents Respond to Imperfect Typhoon Warnings in a Multi-Echelon Supply Chain

**Hanseon Joo[1], Minjong Cheon[2], and Ook Lee[1,*]**

[1] Department of Information Systems, Hanyang University, 222 Wangsimni-ro, Seongdong-gu, Seoul 04763, Republic of Korea

[2] Department of Computer Science, Korea National Open University, 86 Daehak-ro, Jongno-gu, Seoul 03087, Republic of Korea

[*] Corresponding author: Ook Lee (ooklee@hanyang.ac.kr)

E-mail addresses: Hanseon Joo, jooking012@hanyang.ac.kr; Minjong Cheon, jmj2316@mail.knou.ac.kr; Ook Lee, ooklee@hanyang.ac.kr

## ABSTRACT

Large language model (LLM) agents are increasingly studied as supply chain ordering agents. However, it is not known how these agents respond to a probabilistic warning of an exceptional event such as a typhoon. We test them in a four-echelon beer-game supply chain with an import port, under a typhoon warning revised as the storm approaches. Port suspension and recovery are based on a statistical analysis of past typhoon tracks and daily ship arrivals. An order-up-to policy and an ordering rule estimated from human beer-game players serve as the benchmarks. In the benchmarks, the importer gains a little by stocking up slightly. Facing the same warnings, however, the LLM agents order up to about four times as many extra units as a benchmark chain whose buffer equals the expected shortage, and most stock up even at the retailer and wholesaler, which the port does not supply directly. On average the warning raises the cost of every LLM chain, while it lowers the cost of the best benchmark. The rise is significant for two of the four models, whose chains pay more on every demand path. Even when the warning proves right and the port suspends, it does not measurably lower the cost of any LLM chain. By testing and comparing LLM ordering agents in a simulated typhoon, this study shows that a warning can raise the cost of a supply chain they run.

## I. INTRODUCTION

When a typhoon approaches, ports close for days, and clearing the queues of cargo they leave behind takes days to weeks, which disrupts the supply chains that pass through them [1], [2]. Unlike most supply disruptions, a typhoon can be seen coming, because satellites observe the storm over the ocean and global weather models predict its track several days ahead [3]. Forecasters issue warnings several days before the storm arrives and revise them as the track becomes clearer [4], [5]. A typhoon warning, however, does not mean that the storm will certainly arrive. It states a probability, and that probability keeps changing as the storm approaches. The Korea Meteorological Administration (KMA) and the Japan Meteorological Agency (JMA) draw each forecast position as a circle that the storm center enters with a probability of 70%, so the storm passes outside it 30% of the time [4], [5]. Firms that import through the port must therefore place their orders before they know whether the port will suspend operations.

A warning gives firms time to prepare, but it helps only if they prepare in proportion to the risk. If they respond too much and the port does not suspend operations but stays open, too much extra stock remains in the chain, and the extra orders are amplified upstream into a bullwhip effect [6]. How decision makers actually respond to information about a disruption has been studied in beer-game experiments [7], [8]. In one of these experiments, Sarkar and Kumar [8] added a supply disruption to the beer game, but the players learned of the disruption only when it began. No one was warned in advance, and no information turned out to be wrong. The question is therefore how ordering decisions respond to a typhoon warning that comes before the disruption and may not come true.

A conventional ordering rule calculates each order with a formula that people have set in advance. Such a rule can use only information given as numbers, such as stock levels or probabilities, and it cannot read information written in words by itself. A large language model (LLM) agent, by contrast, can read information written in words, such as a typhoon warning, and decide for itself how much to change its orders. Several studies have already let LLM agents place orders in supply chains [9]–[11]. However, no study has yet followed how LLM ordering agents respond to a typhoon warning that may not come true while their orders pass through a multi-echelon supply chain, and compared the resulting cost with the cost of conventional ordering policies.

To fill this gap, this study asks three questions, which move from what the agents do, to how their response changes the cost of the supply chain, and finally to which ordering behaviors go with the size of that change.

- **RQ1 (reaction).** How do LLM ordering agents respond to typhoon warnings?
- **RQ2 (value).** Does the warning system lower the expected cost of a supply chain run by LLM agents?
- **RQ3 (mechanism).** Which ordering behaviors go with a larger change in the cost of a supply chain run by LLM agents?

We answer these three questions in a typhoon version of the beer game, which is the four-echelon serial supply chain widely used in behavioral operations research. Our supply chain consists of an overseas factory, an importer, a wholesaler and a retailer, and a port sits between the overseas factory and the importer. The port and the typhoon in our experiment are not real but statistical, and how often the port suspends operations, when the suspension starts and how fast the port recovers follow daily port calls at ports around the world during past typhoons. For brevity, this paper uses typhoon for any tropical cyclone with maximum sustained winds of at least 34 kt (about 17 m/s), wherever it forms. Each company is run either by an LLM agent or by a conventional ordering policy, and all four companies receive the same typhoon warnings, each of which turns the typhoon forecast into the probability that the port will suspend operations. We run each supply chain with and without the warnings and compare their expected costs, weighting the cost of each possible course of the typhoon by its probability.

This study makes two contributions.

1. **An ordering experiment in a multi-echelon supply chain in which a warning is issued before a disruption and is then updated.** To our knowledge, no earlier ordering experiment has given decision makers a typhoon warning that is issued days before a port suspension, is updated once and may not come true even after the update, while their orders pass through a multi-echelon supply chain. Benchmark ordering policies that receive the same warnings provide a reference for judging whether a response to the warnings is too large or too small. The same design can be used to test other decision makers, such as human players or other ordering agents, under the same warnings.

2. **A chain-wide test of how LLM ordering agents respond to typhoon warnings and how much their response costs the supply chain.** LLM agents place the orders at every company in the chain every day, from the first warning until the port has recovered, and we follow each of these orders. We compare their responses and costs with those of the benchmark policies. We also examine which ordering behaviors go with a larger change in the cost of the chain.

## II. RELATED WORK

### A. The Beer Game and LLM Ordering Agents

The beer game grew out of the industrial dynamics of Forrester [12]. It shows that delays in orders and shipments can turn a small change in customer demand into large swings in orders toward the factory. Sterman [13] reproduced these swings in a controlled beer-game experiment. He showed that the swings arise because players, when placing a new order, do not fully count the supply line, the quantity ordered but not yet received, and therefore keep ordering. Lee et al. [6] analyzed these swings, known as the bullwhip effect, and traced them to four operational causes, namely demand forecasting, order batching, rationing and price fluctuations. Croson and Donohue [14] removed these four operational causes of the bullwhip effect, so that every player knew the distribution of customer demand, orders did not need to be batched, no stock had to be rationed and prices stayed constant. The bullwhip effect still remained, and they traced it in part to the same cause, the players' failure to fully count the supply line.

Several studies let LLM agents place inventory orders in supply chains [9], [15]–[18]. Other studies tested whether LLMs show, in operations decisions such as inventory ordering, the same decision biases that people show [10], [19], [20]. Like the human players in the studies above, LLM agents fail to fully count the supply line, which contributes to the bullwhip effect [10], and agents that negotiate with each other can reduce the bullwhip effect [17]. Long et al. [11] found that the orders of LLM agents varied widely when the same beer game was run again, and they added guardrails, such as caps on the budget, to keep the agents from over-ordering in panic. Dhar [21] simulated a supply chain run by generative AI agents under several kinds of disruption and proposed, but did not test, an agent that watches for early signs of a disruption.

### B. Supply Disruptions in the Beer Game and at Typhoon-Hit Ports

Rong et al. [7] added random drops in the factory's capacity to the beer game. Some players over-reacted to these drops, so that order variability grew toward the retailer, the reverse of the usual bullwhip effect. In a beer game in which the retailer or the factory stopped operating for five periods, Sarkar and Kumar [8] found that sharing information about the factory's stoppage when it began lowered order variability and cost. Cui et al. [22] gave LLM buyers in a simulated drug supply chain shortage alerts before a shortage was confirmed, and these advance alerts more than doubled stockpiling, while the total supply stayed largely unchanged.

Other studies examine how supply chains prepare for and respond to port closures and typhoons. Taskin and Lodree [23] built a Bayesian model that updates hurricane forecasts and decides how much emergency supply to stock. Lewis et al. [24] modeled a port of entry that can close and showed that, when the congestion after a closure matters, the best ordering policy may depend on the state of the port. Verschuur et al. [1] used vessel tracking data to estimate how long natural disasters disrupt ports and found that the disruptions lasted six days at the median. Kuai et al. [25] built a benchmark in which multimodal LLMs read hurricane forecast maps and written forecasts and answered questions about how a port would be affected and which alert level it should set. The models understood the forecasts fairly well but often set a lower alert level than the situation required.

Earlier studies have shown how people and LLM agents order once a disruption begins or is announced. They have also shown how ordering policies should respond to port closures and hurricane forecasts, and how LLMs read a forecast to make a single decision. This study goes a step further than these studies, including Cui et al. [22], and follows how LLM ordering agents in a four-echelon supply chain respond, day after day, to an updated typhoon warning that may not come true.

## III. SIMULATION ENVIRONMENT

### A. Supply Chain and Demand

As in the beer game of Sterman [13], four echelons form a serial chain from a retailer through a wholesaler and an importer to an overseas factory whose production capacity is unlimited. A port lies between the factory and the importer, so a suspension stops the importer's inbound shipments. The wholesaler buys from the importer and the retailer buys from the wholesaler, so both keep receiving goods during a suspension as long as the echelons above them still have stock.

As in Croson and Donohue [14], consumer demand is drawn independently each day from the whole numbers 0 to 8 with equal probability, and mean daily demand is $\mu = 4$ units. One period is one day, because a port suspension lasts days rather than weeks, and inventory models of port closures also use daily periods [24]. The order and shipping delays of two periods, the production delay of three periods and the initial stock of 12 units also follow Croson and Donohue [14], and the holding and backlog costs of 0.5 and 1.0 per unit follow Sterman [13] (Table I). The run lasts 40 days, but the agents are told that the business runs about 60 days, as Sterman [13] did to avoid end-of-game effects.

The benchmark policies are run on 1,000 demand paths, and the LLM agents on the first 10 of these paths, on which the two are compared.

**TABLE I**
Base parameters (the full set is in the online repository).

| Element | Setting | Source |
|---|---|---|
| Echelons | Retailer, wholesaler, importer, | Croson and Donohue [14] |

| Element | Setting | Source |
|---|---|---|
| | factory | |
| Consumer demand | Independent uniform whole numbers from 0 to 8 per day (mean 4); the distribution is known to the agents | Croson and Donohue [14] |
| Delays | Order 2 days; shipping 2 days; production 3 days | Croson and Donohue [14] |
| Order of events | Each day, shipments arrive, orders arrive, stock is shipped, costs are charged, and each echelon orders | Sterman [13] |
| Initial state | 12 units on hand; every pipeline slot holds 4 units | Croson and Donohue [14] |
| Costs | Holding 0.5 and backlog 1.0 per unit per day | Sterman [13] |
| Horizon | 40 days simulated; agents are told about 60 days, to avoid end-game effects | Sterman [13] |
| Objective given to agents | Lowest total cost of the whole chain | Jannelli et al. [17] |

### B. Typhoon, Port Suspension and Recovery

The port and the typhoon are not real ones but statistical composites built from past data, and the agents are told only of an "import port". The suspension's probability, timing and recovery come from daily port calls recorded by satellite vessel tracking at 136 ports in 30 countries during 240 typhoons in 2019–2025 (IMF PortWatch [26], [27] matched to IBTrACS [28], [29]). As in earlier studies of port disruptions [1], we use global data so that the port represents ports in general. A disruption is a sharp fall in a port's daily calls around a storm's closest approach (the full method is in the online repository).

**How often.** Of the 1,501 storm–port pairs within 400 km, 46% show a disruption.

**When.** Port calls fall first on the day before the closest approach in 41% of disruptions, more often than on any other day. The simulated typhoon is expected to make landfall on day 18, so the simulated port suspends from 00:00 on day 17, and the authority announces its decision on day 16.

**How long.** Simchi-Levi et al. [30] argue that firms should plan for supply disruptions by their impact rather than by their likelihood, so we set the suspension to three days. In the port data, a disruption costs a median of 1.8 days of normal traffic, and three days lies close to the 90th percentile (3.2 days). Container terminals in Shanghai and Ningbo also stayed closed for about three days during Typhoon

Dolphin in August 2026 [2]. From day 17 to the end of day 19, no cargo passes through the port, and cargo that arrives waits at the port until it reopens on day 20.

**Recovery.** Reopening is not recovery. After reopening, the port has to handle two kinds of cargo, the cargo that keeps arriving at the mean rate of $\mu = 4$ units a day and the cargo that piled up during the suspension. As in the port queue of Lewis et al. [24], the piled-up cargo clears at the port's spare capacity, which we set to $\mu/m$ units a day. Here m is the recovery multiple, the time that the piled-up cargo takes to clear divided by the length of the suspension. The port therefore releases at most $\mu(1 + 1/m)$ units a day, $\mu$ for the new cargo and $\mu/m$ for the piled-up cargo, until the pile is cleared. Over the 488 disruptions that made up half of their lost calls, the median of m is 4, so the port releases at most 4 + 1 = 5 units a day, and the 12 units that pile up during the three-day suspension clear in 4 × 3 = 12 days, by about day 31.

## C. Warning Systems and Scenarios

We compare two warning systems, NONE and WARN, for every chain, both for the benchmark policies and for the LLM agents. In both systems the port may or may not be suspended, and it is suspended with the same probability of 0.5, but under NONE the agents receive no warning before the authority's decision. NONE is the baseline, so comparing the two systems measures what the warning itself is worth.

- **NONE.** A typhoon approaches but no warnings are issued. The agents learn of a suspension only from the authority's notice on day 16, and no notice is issued if the port stays open. In both systems, the notice of a suspension also states that reopening is expected on day 20, and a further notice follows on the day of reopening.
- **WARN.** A first warning on day 11, seven days before the expected landfall, states a 50% suspension probability. On day 15, three days before the expected landfall, the warning is updated. A toward update says that the storm is heading for the port and raises the probability to 70%, and an away update says that it is turning away and lowers it to 30%. On day 16 the authority announces whether the port will be suspended from day 17.
- **Scenarios.** In both systems the port is suspended with probability 0.5, the rate of 46% from past typhoons (Section III-B) rounded up, and the first warning states this probability. The update on day 15 points the right way 70% of the time, so a toward update comes before 70% of suspensions and an away update before 70% of the cases in which the port stays open. We set this 70% by analogy with the 70% probability radius that the weather agencies of Korea and Japan draw around each forecast position, including the position three days ahead [4], [5]. The four WARN scenarios thus have probabilities 0.5 × 0.7 = 0.35 for a toward update followed by a suspension, 0.15 for a toward update followed by an open port, 0.15 for an away update followed by a suspension and 0.35 for an away update followed by an open port. Each update therefore occurs half of the time, and after a toward update the port is suspended with probability 0.35 / 0.5 = 0.7, so each update states the true chance of a suspension. NONE has two scenarios, suspended or open, each with probability 0.5, and with the four of WARN this gives six scenarios. Each system is evaluated by its expected cost, the average of its scenario costs weighted by their probabilities, so the comparison includes both the cases in which the port is suspended and those in which it stays open.

Because an order arrives four days after it is placed (Table I), only the importer's orders placed on day 12 or earlier arrive before the suspension starts on day 17. A warning that is to help build stock must therefore come before day 13, and the first warning on day 11 leaves two days, day 11 and day 12, to order. Seven days lies beyond the official five-day track forecasts [5], but forecasts of that range are already in use. The US National Hurricane Center extended its outlook of tropical cyclone formation from five to seven days in 2023 [31], the forecasts of the European Centre for Medium-Range Weather Forecasts track tropical cyclones up to ten days ahead [32], and the AI weather model KARINA exceeds the global forecast skill of that center's numerical model up to ten days ahead [33]. Three days ahead the forecast is much more accurate, with track errors smaller than the radius of gale-force winds, so the update can say whether the storm is heading for the port or turning away [34], [35]. The update on day 15 comes too late to build stock before the suspension, but it lets the agents decide whether to stop the extra orders that they began after the first warning.

## IV. ORDERING AGENTS AND EVALUATION

### A. Benchmark Policies and the Expected-Value Buffer

The order-up-to policy orders, each day, the amount that brings its inventory position back to a fixed target stock. The inventory position is the stock on hand, minus the backlog, plus the supply line. The target stock is the demand expected until today's order arrives, plus a small safety stock. For the retailer, the wholesaler and the importer, an order arrives after four days (two days of order delay and two days of shipping), so their expected demand is 4 × 4 = 16 units. The factory refills its stock by production, which takes three days, so its expected demand is 4 × 3 = 12 units. A grid search on the no-typhoon scenario over 400 separate demand paths finds target stocks of 20 for these three echelons and 14 for the factory to be the cheapest, so the safety stock is 4 units for these three echelons and 2 units for the factory. The safety stock is small because a unit of backlog costs 1.0 a day and a unit held in stock costs 0.5 a day, so it is cheapest to accept a shortage with a probability of about one third (the derivation is in the online repository). This rule gives a safety stock of 2 units at every echelon, but the grid search finds 4 units for the retailer, the wholesaler and the importer, because their supplier sometimes runs short and delivers late. These target stocks are the best ones for normal weather, not for a typhoon.

When a supply disruption is announced in advance, optimal inventory policies raise the order-up-to level ahead of the disruption [36]. We use the simplest such rule as a reference for the size of a reaction. A suspension of D = 3 days cuts off the importer's inbound supply, so the importer loses about $D \cdot \mu = 12$ units. While a warning of probability p is in force, the expected-value response raises the importer's target stock by its expected loss $p \cdot D \cdot \mu$, which is 0.5 × 12 = 6 units under a 50% warning and 0.7 × 12 = 8.4 units under a 70% warning. We multiply every response by a reaction multiplier k, so that on day t echelon i adds the following buffer to its target stock.

$$\Delta_{i,t} = k\, p_t\, \mathrm{D}\, \mu \cdot \mathbb{1}[i \in \mathscr{S}]$$

Here $p_t$ is the probability in the warning on day t, $\mathscr{S}$ is the set of echelons that stock up, and $\mathbb{1}[i \in \mathscr{S}]$ is 1 if echelon i is in this set and 0 otherwise. The value k = 1 is the expected-value buffer, k < 1 is a smaller response and k > 1 is an over-reaction. We test two scopes, one in which only the importer, which receives its supply through the port, stocks up, and one in which all four echelons stock up by the same rule. The buffer follows the probability in the warning, including the update on day 15, is kept at its last size from the day on which a suspension is announced until the port reopens, and is removed when the port reopens or when the typhoon passes without a suspension. We test k = 0, 0.25, 0.5, 0.75, 1, 1.25, 1.5, 2, 3 and 4, where k = 0 ignores the warning.

The human-like heuristic uses the anchoring-and-adjustment rule of Sterman [13] with his mean estimates for human subjects. Sterman estimated these values in a weekly game with different demand, so we use the rule as a simple ordering rule of the kind people follow, not as an exact model of people in this chain. It does not describe how people react to warnings, so we give it the same buffer rule as the order-up-to policy. Table II summarizes the benchmark policies.

**TABLE II**
Benchmark policies.

| Policy | Rule | Role |
|---|---|---|
| Order-up-to, warning-blind | Order up to a fixed inventory position of 20 for the retailer, the wholesaler and the importer and 14 for the factory | A cost-efficient reference that ignores warnings |
| Order-up-to with warning buffer | As above, plus $\Delta = k \cdot p \cdot D \cdot \mu$ at the importer or at every echelon; the buffer is held from the announcement of a suspension until reopening | Responses of size k, including the expected-value buffer (k = 1) |
| Human-like heuristic | Anchoring and adjustment with Sterman's [13] mean estimates $\theta = 0.36$, $\alpha_S = 0.26$, $\beta = 0.34$, $S' = 17$, with or without the buffer | A proxy for how people order |

## B. LLM Agents

All four echelons are run by the same model, but each echelon is a separate agent that makes its own decisions, so that differences between chains reflect differences between models. Every day an agent sees its role, its supplier and its customer, the lead times and costs, and the demand distribution. It also sees today's receipts, incoming order, stock, backlog, supply line and port status. Below these it sees its own record of the run, one line for each past day, and all warnings and notices issued so far, which are the same for all four companies. It does not see the stock, orders or records of the other three companies. The prompt never mentions the beer game, to limit recall of a well-known exercise. Each day, the agent replies in JSON with its order, a whole number, and a one-sentence reason for that order (Appendix A). We read the order from the first JSON object in a reply. In 11 of the 6,640 Sonnet replies, the agent wrote a second JSON object with an order 1 to 3 units lower, and we still used the first order. All 11 came from the importer after the port reopened.

Jannelli et al. [17] compared models of different size from one provider, and we widen the comparison to two models from each of two providers, Claude Sonnet 5.5 and Claude Opus 5.5 from Anthropic and GPT-6.1 Terra and GPT-6.2 Sol from OpenAI. All four models ran at the provider's default sampling, without a temperature setting, and with the reasoning effort set to low. With this setting, Opus and Terra reasoned before most replies, using on average 146 and 51 reasoning tokens per reply.

**Prefix sharing.** For each model and demand path, the six scenarios give an agent exactly the same prompt until the warnings or notices that they show differ. When an agent on a given demand path faces a prompt that it has already answered, the recorded answer is reused. An answer is therefore reused only when the model, the demand path, the day, the echelon and every word of the prompt are the same. An LLM can give a different answer when it is asked the same prompt again. Without sharing, the scenarios could differ by chance before any warning. Sharing the answers keeps the six scenarios of one model and one demand path identical until their prompts differ, so on the day the warnings or notices first differ, the prompts differ only in the warning or notice, while the stock, the orders and the record are still the same. This is common random numbers applied to decisions. From that day on, each scenario asks the model separately, so the differences between scenarios also include some chance variation in the answers.

## C. Value of a Warning System and Other Measures

Following the economic value of forecasts [37], we judge a warning system by what it does to the expected cost of the whole chain for a given decision maker. Let W be the warning system and N the absence of one. The value of the warning system is the expected cost without it minus the expected cost with it,

$$V_W = \mathrm{E}[C \mid N] - \mathrm{E}[C \mid W]$$

As in Section III-C, the expectation weights the cost of each scenario by its probability, and $V_W$ splits into two parts. The first part is the gain in the courses where the port suspends. It is positive if the agents prepared well. The second part is the loss in the courses where the port stays open. It is positive if the agents stocked up for nothing. A decision maker who optimizes can always ignore an unhelpful warning, so for that decision maker $V \geq 0$. A boundedly rational agent can make V negative by reacting too much, too widely or too late. We compute V on each demand path and then average it, so every comparison is paired.

- **Chain cost.** The chain cost is the total holding and backlog cost of all four companies over 40 days, and we report its expectation under each system.
- **Panic ratio.** When V is negative, the panic ratio is $-V$ divided by the expected cost of the typhoon to the order-up-to chain without a warning on the same demand paths. The expected cost of the typhoon is half of the difference between this chain's cost when the port suspends and its cost when the port stays open, both without a warning. We divide every chain by the same cost, so that the ratios can be compared across chains. A ratio above 1 means that the response to the warning costs more than the typhoon costs a well-run chain.
- **Reaction index RI.** For each chain, we add up the orders of all four echelons in the warning window (days 11 to 15) and subtract the same sum in the chain's NONE run on the same path. RI is this excess divided by the excess of the order-up-to benchmark with the expected-value buffer ($k = 1$). RI = 1 means that the chain ordered as much extra stock as the expected-value buffer, and RI = 2 means twice as much. Each chain can therefore be compared with order-up-to buffers of the same RI, at the importer only or at every echelon. We also report the excess orders of each echelon.

- **Updating.** We measure the excess orders on the day of an away update and on the day of a toward update, and the surplus stock in the 12 days after the authority announces that the port stays open. Surplus stock is measured in unit-days of stock above the same agent's NONE run.
- **Bullwhip ratio.** The bullwhip ratio is the variance of an echelon's daily orders divided by the variance of consumer demand, and we report it echelon by echelon. A ratio above 1 means that the echelon's orders swing more than consumer demand, and a ratio that rises toward the factory means amplification along the chain.
- **Ordering regression.** We regress the daily order on lagged net stock, the incoming order, receipts, the supply line and time [8, eq. 1], plus the stated probability, a port-suspension indicator and the agent's own order on the previous day, by echelon and model. Each decision enters the regression once, so the days that the six scenarios of a demand path share are not counted more than once. The intervals come from 2,000 bootstrap resamples of demand paths. This shows whether warnings raise orders directly, beyond what the gap in stock implies, and whether the agents count their supply line. The previous order captures how smoothly an agent orders, so that this smoothing is not mistaken for counting the supply line.

### D. Experimental Runs and Statistical Tests

**One typhoon for every path.** Every demand path and every scenario face the same typhoon. The intervals we report therefore reflect variation in demand, not in the typhoon.

Every comparison is made between runs on the same demand path. We report 95% bootstrap intervals over demand paths (the 10 paths of the LLM experiment are drawn again 5,000 times, and the 1,000 paths of each benchmark result 2,000 times). When an interval does not include zero, the variation between demand paths alone is unlikely to explain the result. Each decision of an LLM agent is drawn once, so the intervals of the LLM chains contain the variation in the models' answers together with the variation between demand paths, and the two cannot be separated. For each LLM chain, we compare its value of the warning with that of the benchmark with the highest value of the warning on these 10 paths (order-up-to, $k = 0.5$), path by path, and we test with a two-sided Wilcoxon signed-rank test whether the difference between them is larger than chance would explain. Because we test four LLM chains, we apply the Holm correction. We also use the same Wilcoxon test to check whether the value of the warning for each LLM chain differs from zero. With only 10 demand paths, the p-value of this test cannot be smaller than 0.002, and it reaches this value only when a chain does worse on all 10 paths.

## V. BENCHMARK RESULTS

### A. Validation

The human-like heuristic reproduces the upstream amplification found in human chains [13], [14]. Without a typhoon, its bullwhip ratios rise from 0.75 at the retailer to 1.32 at the wholesaler, 2.48 at the importer and 3.76 at the factory. The order-up-to policy passes demand on almost unchanged, with ratios from 1.06 to 1.19.

Without a warning system, the suspension raises the order-up-to chain's cost from 593.8 on the paths where the port stays open to 709.3 on the paths where it suspends, an increase of 19.4%. The expected cost of the typhoon is half of this difference, 57.7. The order-up-to chain is lean. Its safety stock is only four units at each downstream echelon, which is one day of demand, and the rest of its target is in transit. A three-day stop of inbound supply therefore turns into backlog all the way to the retailer. For the heuristic chain, the suspension adds only 33.8 (from 1,336.2 to 1,370.1, +2.5%), because the heuristic already holds much more stock and its own bullwhip effect drives most of its cost.

## B. Reaction Size and Scope

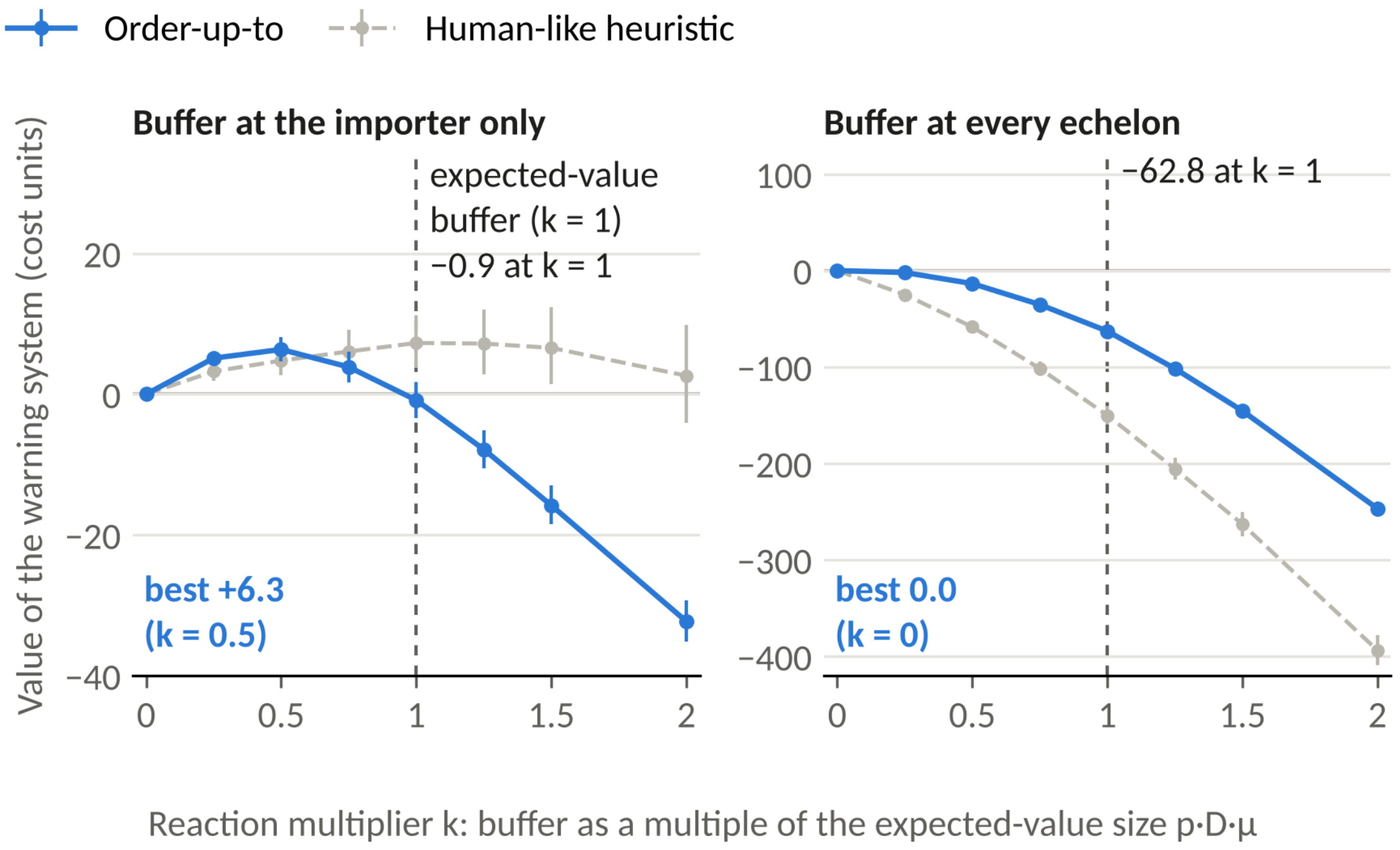

Fig. 1. Value of the warning system against the reaction multiplier k, with the buffer at the importer only (left) or at every echelon (right), for the order-up-to chain and the human-like heuristic. The panels have different scales. Above the zero line the warning lowers expected cost, and k = 1 is the expected-value buffer. The figure shows k up to 2, and the text gives the values for larger k.

With the buffer at the importer only (Fig. 1, left panel), the value of the warning is highest at k = 0.5, where it is 6.3. It falls to about zero at the expected-value buffer (−0.9 at k = 1) and becomes clearly negative beyond it, reaching −110.9 at k = 4. The split into gain and loss shows why. At k = 0.5 the paths where the port suspends gain 8.5 and the paths where it stays open lose 2.1. At k = 1 the gain falls to 7.0 while the loss rises to 7.8. At k = 1.5 the gain when the port suspends falls to zero (−0.04), and beyond it the buffer loses money even when the port suspends (−7.5 at k = 2 and −41.4 at k = 4).

A full buffer helps less than its size suggests, even when the port suspends. On the path with a toward update and a suspension, the expected-value buffer cuts backlog at the retailer, the wholesaler and the importer from 254.9 to 213.0 unit-days, a saving of 41.9. Against this, it adds 16.0 in holding cost at those three echelons and 13.4 in cost at the factory, mostly backlog (8.9), because the factory cannot fill the importer's extra orders at once. The path therefore costs 696.8 rather than 709.3. The half-size buffer reaches 692.8 with less holding. On the path with a toward update after which the port stays open, the full buffer costs 18.0 more than no buffer (611.8 against 593.8), mostly as holding cost at the importer and as backlog at the factory.

When every echelon stocks up (Fig. 1, right panel), the warning does not lower cost at any reaction size, with −2.0 at k = 0.25, −62.8 at k = 1 and −787.6 at k = 4. The downstream echelons pull the importer's stock away before the suspension starts. On the path with a toward update and a suspension at k = 1, the retailer's backlog falls from 65.7 to 38.0 unit-days, but the importer's rises from 107.9 to 146.6, and holding at the retailer and at the factory adds 65.1. The path costs 786.9, which is 77.6 more than without a warning. Stocking up downstream moves the buffer away from the echelon that sits behind the port. In terms of the panic ratio, an importer buffer four times the expected-value size costs the order-up-to chain 1.9 times the typhoon's expected cost (110.9 against 57.7), and the same reaction at every echelon costs 13.6 times.

The human-like heuristic gains most at a larger multiplier than the order-up-to chain (k = 1 to 1.25 rather than k = 0.5). Its value rises to 7.3 at k = 1, stays positive up to k = 2 (2.6) and falls to −45.1 at k = 4. The reason is that the heuristic closes a stock gap slowly ($\alpha_S$ = 0.26), so only part of a raised target is ordered while the warning is in force. When every echelon stocks up, the heuristic loses 150.1 at k = 1 and 1,048.9 at k = 4.

## C. The Demand Paths of the LLM Experiment

On the 10 demand paths that the LLM agents face, the benchmarks reproduce the pattern of the 1,000 paths (Table III). On these ten paths the suspension is cheaper than on average (it adds 55.1 rather than 115.5), but the importer-only buffers with k = 0.5 or more gain slightly more here than over the 1,000 paths.

**TABLE III**

Value of the warning system on the 1,000 benchmark paths and on the 10 paths of the LLM experiment (cost units per run, paired 95% bootstrap intervals).

| Policy | Buffer at | k | 1,000 paths | 10 LLM paths |
|---|---|---|---|---|
| Order-up-to | Importer | 0.25 | 5.1 [4.1, 6.1] | 4.1 [−2.3, 11.5] |
| Order-up-to | Importer | 0.5 | 6.3 [4.6, 8.1] | 7.8 [−6.9, 26.6] |
| Order-up-to | Importer | 0.75 | 3.9 [1.6, 6.1] | 6.2 [−13.9, 34.5] |
| Order-up-to | Importer | 1 | −0.9 [−3.4, 1.7] | 2.0 [−20.8, 35.0] |
| Order-up-to | Importer | 1.25 | −7.8 [−10.5, −5.2] | −3.1 [−29.5, 35.9] |
| Order-up-to | Importer | 1.5 | −15.8 [−18.5, −13.0] | −9.2 [−38.7, 33.6] |
| Order-up-to | Importer | 2 | −32.3 [−35.1, −29.3] | −25.7 [−57.9, 20.0] |
| Order-up-to | Importer | 3 | −68.5 [−71.5, −65.4] | −61.8 [−98.8, −12.1] |
| Order-up-to | Importer | 4 | −110.9 [−114.3, −107.5] | −106.0 [−147.4, −52.1] |

| Policy | Buffer at | k | 1,000 paths | 10 LLM paths |
|---|---|---|---|---|
| | | | | |
| Order-up-to | Every echelon | 1 | −62.8 [−67.3, −58.3] | −72.2 [−113.4, −19.5] |
| Order-up-to | Every echelon | 4 | −787.6 [−798.1, −777.4] | −811.1 [−910.9, −684.6] |
| Human-like heuristic | Importer | 1 | 7.3 [3.7, 11.2] | 6.3 [−21.0, 36.1] |
| Human-like heuristic | Every echelon | 4 | −1,048.9 [−1,075.5, −1,021.6] | −971.1 [−1,239.9, −677.5] |

### D. Sensitivity

We repeated the benchmark runs on the same 1,000 demand paths in eight other settings, changing one setting at a time (the full table is in the online repository). The port recovers faster or slower (m = 2 or 8), the suspension lasts two days, each path draws its typhoon from the 683 disruptions worldwide, the update is more accurate (80% or 90% after a toward update and 20% or 10% after an away update), the backlog cost is doubled, or shipping takes three days. When a setting changes the costs or the delays, the order-up-to levels are tuned again.

In every setting, the best buffer at the importer is a quarter or half of the expected-value buffer (k = 0.25 or 0.5) and gains 1.8 to 14.2. An importer buffer twice the expected-value size loses 19.3 to 42.0, and an expected-value buffer at every echelon loses 35.4 to 95.9. The expected-value buffer at the importer gains in three settings, the slower recovery, the doubled backlog cost and the longer shipping delay (0.8 to 7.0), and loses in the other five and in the base case (0.9 to 8.5).

## VI. LLM RESULTS

The warning affects the four LLM chains very differently (Table IV). It raises the cost of the Terra chain by 17.0 per run and that of the Sol chain by 26.3, and their cost with the warning is higher than without it on only 5 and 7 of the 10 paths. It raises the cost of the Opus chain by 61.9 and that of the Sonnet chain by 169.4, and their cost with the warning is higher on all 10 paths.

In normal weather, the Terra, Sol, Opus and Sonnet chains cost 4%, 16%, 44% and 77% more than the order-up-to chain, mainly because the Opus and Sonnet chains hold more stock. Without a warning, the suspension adds 47.3 to the cost of the Terra chain, 37.2 to the Sol chain, 15.0 to the Opus chain and 14.0 to the Sonnet chain, against 55.1 for the order-up-to chain. Even when the port suspends, the Opus and Sonnet chains still cost far more in total than the order-up-to chain.

**TABLE IV**
The six chains on the 10 demand paths (cost units per run). The last two columns show how much the warning changes the cost on a toward path that ends in a suspension and on an away path after which the port stays open, and a negative change means that the warning lowers cost.

| Chain | Cost in normal weather (no warning, port open) | Value of the warning V | Change in cost from the warning when the port suspends | Change in cost from the warning when the port stays open |
|---|---|---|---|---|
| Order-up-to, k = 1 | 593.4 | +2.0 | −25.3 | +20.4 |
| Human-like heuristic, k = 1 | 1,284.2 | +6.3 | −12.1 | −1.8 |
| Sonnet 5.5 | 1,051.1 | −169.4 | +204.0 | +144.1 |
| Opus 5.5 | 851.9 | −61.9 | +67.9 | +55.3 |
| GPT-6.1 Terra | 618.2 | −17.0 | +10.1 | +28.2 |
| GPT-6.2 Sol | 686.1 | −26.3 | +5.6 | +43.0 |

## A. Reaction to Warnings (RQ1)

Until the authority's decision on day 16, the paths that end in a suspension and the paths on which the port stays open are identical, so the warning window has only two cases per model, a toward update and an away update.

Three of the four LLM chains over-react (Table V). On the toward path, the expected-value benchmark orders 14.0 extra units in the warning window, against 12.5 for Terra, 25.7 for Sol, 46.8 for Opus and 33.1 for Sonnet. The reaction index RI (Section IV-C) is therefore 0.89 for Terra, 1.84 for Sol, 3.34 for Opus and 2.36 for Sonnet. Terra orders about as much as the expected-value benchmark, whose RI is 1, and the Sol, Opus and Sonnet chains order more than this benchmark on all 10 paths.

In the expected-value benchmark, only the importer adds a buffer. Its factory orders 6.0 extra units only because it passes on the importer's extra orders, and its retailer and wholesaler order nothing extra. In the Terra chain, the retailer and the wholesaler order only 0.5 to 1.1 extra units, while in the Sol, Opus and Sonnet chains they order 2.4 to 9.9 extra units.

In the warning window, the share of the reasons that the agents wrote with their orders and that name the typhoon, the port, the suspension or a closure is 97% at the Terra importer but only 49% to 54% at the other three Terra echelons, 74% to 97% at the Sol echelons, and 100% at every Opus and Sonnet echelon. A typical reason, written by the Sonnet retailer on day 11, reads "a possible port suspension around day 17-20 argues for ordering slightly above the mean demand of 4 to build a small buffer".

**TABLE V**

Reaction index (RI) and extra orders in the warning window (days 11 to 15), by echelon, compared with the same chain without a warning.

| Chain | Warning path | RI, chain [95% CI] | Excess, retailer | Excess, wholesaler | Excess, importer | Excess, factory |
|---|---|---|---|---|---|---|
| Order-up-to, k = 1 | Toward update | 1 | 0 | 0 | 8.0 | 6.0 |
| Order-up-to, k = 1 | Away update | 1 | 0 | 0 | 4.1 | 6.0 |
| Sonnet 5.5 | Toward update | 2.36 [2.05, 2.67] | 5.2 | 7.2 | 10.6 | 10.1 |
| Sonnet 5.5 | Away update | 2.70 [2.33, 3.08] | 4.0 | 5.6 | 9.0 | 8.7 |
| Opus 5.5 | Toward update | 3.34 [3.05, 3.64] | 5.8 | 9.9 | 15.5 | 15.6 |
| Opus 5.5 | Away update | 3.86 [3.48, 4.26] | 4.7 | 8.4 | 11.8 | 14.0 |
| GPT-6.1 Terra | Toward update | 0.89 [0.61, 1.19] | 0.6 | 1.1 | 4.0 | 6.8 |
| GPT-6.1 Terra | Away update | 1.13 [0.73, 1.47] | 0.5 | 0.6 | 3.5 | 6.8 |
| GPT-6.2 Sol | Toward update | 1.84 [1.51, 2.19] | 3.2 | 4.5 | 9.2 | 8.8 |
| GPT-6.2 Sol | Away update | 2.30 [1.94, 2.71] | 2.4 | 3.8 | 8.6 | 8.5 |

## B. Response to the Update (RQ1)

On average, all four LLM chains order more after a toward update than after an away update (Table VI), although for Terra this holds on only 5 of the 10 paths. On the day of the update, the difference between the two updates is 1.1 units for Terra, 2.4 for Sol, 7.9 for Opus and 5.8 for Sonnet, against 3.9 for the expected-value benchmark.

After an away update, the benchmark orders 1.9 units less than normal and so sheds part of its buffer. The Terra chain orders about as much as normal (−0.1), while the Sol, Opus and Sonnet chains still order more than normal (0.6, 1.6 and 5.4 units).

Most of the extra stock was ordered earlier, under the 50% warning, and orders cannot be canceled. In the 12 days after the port is declared open, the LLM chains therefore carry 1.3 to 3.8 times the surplus stock of the expected-value benchmark.

**TABLE VI**
Extra orders under the first warning (days 11 to 14) and on the day of the update (day 15), and surplus stock in the 12 days after the port is declared open, summed over the chain.

| Chain | Built under the 50% warning | Orders after an away update | Orders after a toward update | Surplus stock, away path | Surplus stock, toward path |
|---|---|---|---|---|---|
| Order-up-to, k = 1 | 12.0 | −1.9 | 2.0 | 36.4 | 46.5 |
| Order-up-to, k = 0.5 | 6.0 | −1.0 | 1.0 | 18.6 | 21.7 |
| Human-like heuristic, k = 1 | 6.1 | 1.4 | 2.5 | 37.1 | 43.3 |
| Sonnet 5.5 | 21.9 | 5.4 | 11.2 | 134.4 | 142.1 |
| Opus 5.5 | 37.3 | 1.6 | 9.5 | 138.7 | 147.1 |
| GPT-6.1 Terra | 11.5 | −0.1 | 1.0 | 54.4 | 62.7 |
| GPT-6.2 Sol | 22.7 | 0.6 | 3.0 | 104.1 | 105.2 |

## C. Value of the Warning (RQ2)

Table VII reports the value of the warning. The benchmarks that react to the warning gain a little from it on average (2.0 to 7.8). For every LLM chain the value is negative. It is −17.0 for Terra, −26.3 for Sol, −61.9 for Opus and −169.4 for Sonnet. In a Wilcoxon signed-rank test, the value differs from zero for Opus and Sonnet ($p = 0.002$ for each) but not for Terra ($p = 0.56$) or Sol ($p = 0.19$).

The part of V that comes from the runs in which the port suspends is −2.8 for Terra, −2.7 for Sol, −33.0 for Opus and −94.8 for Sonnet, so on average no LLM chain gains even when the suspension comes, although Terra gains on 4 of the 10 paths, Sol on 3 and Opus on 1. The part that comes from the runs in which the port stays open is a loss of 14.2, 23.6, 28.8 and 74.6. Of the total loss, 83% for Terra, 90% for Sol, 47% for Opus and 44% for Sonnet comes from these false alarms.

Compared with the best benchmark (order-up-to, k = 0.5), the value of the warning is 24.8 lower for Terra, which does worse on 6 of the 10 paths (Holm-adjusted $p = 0.16$), 34.1 lower for Sol, which does worse on 8 ($p = 0.027$), and 69.7 and 177.2 lower for Opus and Sonnet, which do worse on all 10 ($p = 0.008$ for each). The difference is significant for Sol, Opus and Sonnet, but not for Terra.

The panic ratio (Section IV-C) divides the loss from the warning by the typhoon's expected cost to the order-up-to chain without a warning, which is 27.6 on these paths, half of the suspension cost of 55.1. It is 0.62 for Terra, 0.95 for Sol, 2.25 for Opus and 6.15 for Sonnet, so it exceeds 1 only for Opus and Sonnet. With the typhoon's expected cost over the 1,000 benchmark paths (57.7) as the divisor, the ratio is 0.29, 0.46, 1.07 and 2.93, so it again exceeds 1 only for Opus and Sonnet.

The warning is worth more to every other LLM chain than to Sonnet, by 152.4 for Terra, 143.1 for Sol and 107.5 for Opus, and these chains do better than Sonnet on 10, 10 and 9 of the 10 paths. The values for Terra and Sol differ by 9.3, and Sol is lower on 5 of the 10 paths.

In normal weather, the mean bullwhip ratios of all four LLM chains are below 1 at every echelon. On the path with a toward update and a suspension, the warning keeps the mean ratios of Terra, Sol and Opus below 1 but raises Sonnet's ratios at the importer and the factory to 1.47 and 1.43, so that its orders there swing more than consumer demand, as the importer's orders of the order-up-to benchmark (k = 1) do on this path (1.38).

**TABLE VII**
Value of the warning system V (cost units per run, 10 demand paths). The gain and loss columns are the parts of V from the runs in which the port suspends and those in which it stays open, weighted by the scenario probabilities, so that gain minus loss equals V. V after a toward (away) update is the value given that update. The Holm-adjusted p-value tests the difference from the best benchmark (order-up-to, k = 0.5). The panic ratio divides −V by the typhoon's expected cost on these 10 paths (27.6) and, in parentheses, on the 1,000 benchmark paths (57.7).

| Chain | V [95% CI] | Gain when the port suspends | Loss when it stays open | V after a toward update | V after an away update | Panic ratio | Holm p vs best benchmark |
|---|---|---|---|---|---|---|---|
| Order-up-to, warning-blind | 0 | 0 | 0 | 0 | 0 | — | — |
| Order-up-to, k = 1 | 2.0 [−20.8, 35.0] | 13.0 | 11.0 | 9.9 | −6.0 | — | — |
| Order-up-to, k = 0.5 (best) | 7.8 [−6.9, 26.6] | 12.0 | 4.2 | 14.1 | 1.6 | — | — |
| Human-like heuristic, k = 1 | 6.3 [−21.0, 36.1] | 5.7 | −0.6 | 8.5 | 4.2 | — | — |
| Sonnet 5.5 | −169.4 [−205.2, −135.8] | −94.8 | 74.6 | −191.1 | −147.7 | 6.15 (2.93) | 0.008 |
| Opus 5.5 | −61.9 [−77.5, −45.4] | −33.0 | 28.8 | −66.5 | −57.3 | 2.25 (1.07) | 0.008 |
| GPT-6.1 Terra | −17.0 [−47.2, 9.0] | −2.8 | 14.2 | −15.7 | −18.3 | 0.62 (0.29) | 0.16 |
| GPT-6.2 Sol | −26.3 [−56.7, 7.3] | −2.7 | 23.6 | −21.0 | −31.6 | 0.95 (0.46) | 0.027 |

## D. Ordering Behaviors That Go With the Size of the Loss (RQ3)

The path with a toward update and a suspension shows how the losses arise. On this path, the warning lowers the cost of the order-up-to benchmark (k = 1) by 25.3 but raises it by 10.1 for Terra, 5.6 for Sol, 67.9 for Opus and 204.0 for Sonnet (Table IV). For Terra and Sol, the extra holding cost (42.2 and 54.8) is largely offset by less backlog (−32.1 and −49.2). For Opus, holding cost rises by 78.0, spread over all four echelons, and backlog cost falls by 10.1, almost all of it at the importer after the port reopens, because the Opus wholesaler, which had built up its own stock, then ordered less from the importer.

For Sonnet, the warning adds 130.4 unit-days of backlog to the chain, and almost all of it (127.6) falls on the importer. The Sonnet wholesaler ordered more from the importer than without a warning, both before and during the suspension, and these orders drew stock out of the importer, whose holding cost fell by 36.5 over the run. Because the port was suspended and then released at most 5 units a day, the importer could not receive more than without a warning, and its backlog tripled from 63.7 to 191.3 unit-days, mostly after the port reopened. The losses of Terra, Sol and Opus therefore come from holding extra stock, while most of Sonnet's loss (127.6 of 204.0) comes from this shortage at the importer.

The timing of the extra orders also differs. On the toward path, Terra, Sol and Opus place 51%, 45% and 38% of their extra orders of the warning window on days 11 and 12, when an order can still arrive before the suspension, against 27% for Sonnet. Over days 16 to 19, Terra, Sol and Opus order 4.7, 5.3 and 12.0 units less than in the same suspension without a warning, while Sonnet orders 30.8 units more. Fig. 2 shows these orders day by day, together with the net stock of the importer.

The size of the reaction during the warning (days 11 to 15) does not explain the loss of the Sonnet chain. On the same 10 paths, order-up-to buffers with the same RI on the toward path as each LLM chain (matched by interpolating over the buffer size), placed at the importer only or at every echelon, change the value of the warning by +3.7 and −4.8 at Terra's RI and lose 13.4 to 18.0 at Sol's, 42.1 to 72.6 at Opus's and 23.0 to 36.7 at Sonnet's. Terra loses 17.0, Sol 26.3 and Opus 61.9. Sonnet loses 169.4, 4.6 to 7.4 times as much as buffers with its RI, and does worse than both of them on all 10 paths.

The ordering regression (Table VIII) shows four results. First, the stated probability raises orders significantly at 14 of the 16 echelons.

Second, the net-stock coefficients, which measure how fast an agent closes a gap in its stock, are −0.13 to −0.28 for Terra, Sol and Opus, against −0.26 for the mean of Sterman's human subjects ($-\alpha_S$), but only −0.06 to −0.08 for Sonnet, which closes less than a tenth of a stock gap per day.

Third, relative to the net-stock weight, Terra counts 69% to 90% of its supply line, the stock already ordered but not yet received, Sol 47% to 80% and Opus 52% to 84%, against 34% for Sterman's subjects, and their supply-line coefficients are below zero at every echelon. Sonnet counts 27% at the importer and 13% at the wholesaler, and at the retailer and the factory its coefficient is positive, so there it orders more when more stock is on its way.

Fourth, while the port is suspended, the Sol and Opus importers order less (−0.50 and −0.38), the Terra importer shows no clear change, and the Sonnet importer orders 2.2 units a day more, although these orders can arrive only after reopening.

**TABLE VIII**
Ordering regression by model and echelon (coefficients with 95% bootstrap intervals over demand paths). The dependent variable is the daily order.

| Model | Echelon | Net stock, lagged | Supply line | Stated probability | Port suspended | Own order, lagged |
|---|---|---|---|---|---|---|
| Sonnet 5.5 | Retailer | −0.06 [−0.07, −0.05] | 0.07 [0.05, 0.09] | 1.14 [0.85, 1.50] | 0.48 [0.28, 0.71] | 0.19 [0.13, 0.26] |
| Sonnet 5.5 | Wholesaler | −0.07 [−0.09, −0.05] | −0.01 [−0.02, 0.01] | 1.58 [1.17, 2.09] | 0.96 [0.76, 1.18] | 0.35 [0.30, 0.39] |
| Sonnet 5.5 | Importer | −0.06 [−0.08, −0.05] | −0.02 [−0.03, −0.01] | 2.07 [1.66, 2.55] | 2.22 [1.67, 3.14] | 0.30 [0.23, 0.36] |
| Sonnet 5.5 | Factory | −0.08 [−0.09, −0.07] | 0.05 [0.00, 0.11] | 1.52 [1.28, 1.79] | 0.39 [0.23, 0.58] | 0.25 [0.15, 0.32] |
| Opus 5.5 | Retailer | −0.24 [−0.26, −0.23] | −0.18 [−0.21, −0.15] | 2.73 [2.27, 3.19] | 0.54 [0.36, 0.69] | 0.18 [0.14, 0.22] |
| Opus 5.5 | Wholesaler | −0.19 [−0.23, −0.17] | −0.11 [−0.13, −0.09] | 2.21 [1.84, 2.72] | 0.71 [0.52, 0.95] | 0.18 [0.13, 0.22] |
| Opus 5.5 | Importer | −0.25 [−0.30, −0.21] | −0.21 [−0.25, −0.18] | 4.87 [3.92, 6.02] | −0.38 [−0.56, −0.17] | 0.27 [0.22, 0.31] |
| Opus 5.5 | Factory | −0.18 [−0.25, −0.14] | −0.09 [−0.15, −0.05] | 1.71 [0.98, 2.71] | 0.07 [−0.17, 0.34] | 0.12 [0.02, 0.21] |
| GPT-6.1 Terra | Retailer | −0.28 [−0.33, −0.25] | −0.26 [−0.30, −0.22] | 0.25 [−0.13, 0.59] | 0.07 [−0.04, 0.18] | 0.09 [0.03, 0.15] |
| GPT-6.1 Terra | Wholesaler | −0.13 [−0.19, −0.11] | −0.09 [−0.19, −0.06] | 0.65 [0.45, 0.83] | 0.37 [0.09, 0.66] | 0.12 [0.06, 0.19] |
| GPT-6.1 Terra | Importer | −0.19 [−0.26, −0.15] | −0.16 [−0.23, −0.12] | 1.63 [0.77, 2.51] | −0.30 [−0.58, 0.04] | 0.12 [0.08, 0.16] |
| GPT-6.1 Terra | Factory | −0.22 [−0.30, −0.15] | −0.16 [−0.26, −0.09] | 0.52 [0.05, 0.98] | −0.02 [−0.36, 0.31] | 0.07 [−0.01, 0.15] |
| GPT-6.2 Sol | Retailer | −0.25 [−0.29, −0.21] | −0.20 [−0.25, −0.16] | 1.72 [0.99, 2.45] | 0.17 [−0.09, 0.43] | 0.11 [0.06, 0.16] |
| GPT-6.2 Sol | Wholesaler | −0.16 [−0.21, −0.13] | −0.11 [−0.16, −0.07] | 1.38 [0.90, 1.80] | 0.22 [0.01, 0.48] | 0.15 [0.09, 0.22] |
| GPT-6.2 Sol | Importer | −0.14 [−0.17, −0.11] | −0.09 [−0.11, −0.07] | 2.22 [1.78, 2.69] | −0.50 [−0.72, | 0.18 [0.09, 0.27] |

| Model | Echelon | Net stock, lagged | Supply line | Stated probability | Port suspended | Own order, lagged |
|---|---|---|---|---|---|---|
| | | | | | −0.26] | |
| GPT-6.2 Sol | Factory | −0.15 [−0.20, −0.12] | −0.07 [−0.14, −0.01] | 0.07 [−0.36, 0.60] | 0.09 [−0.05, 0.27] | 0.05 [−0.00, 0.11] |

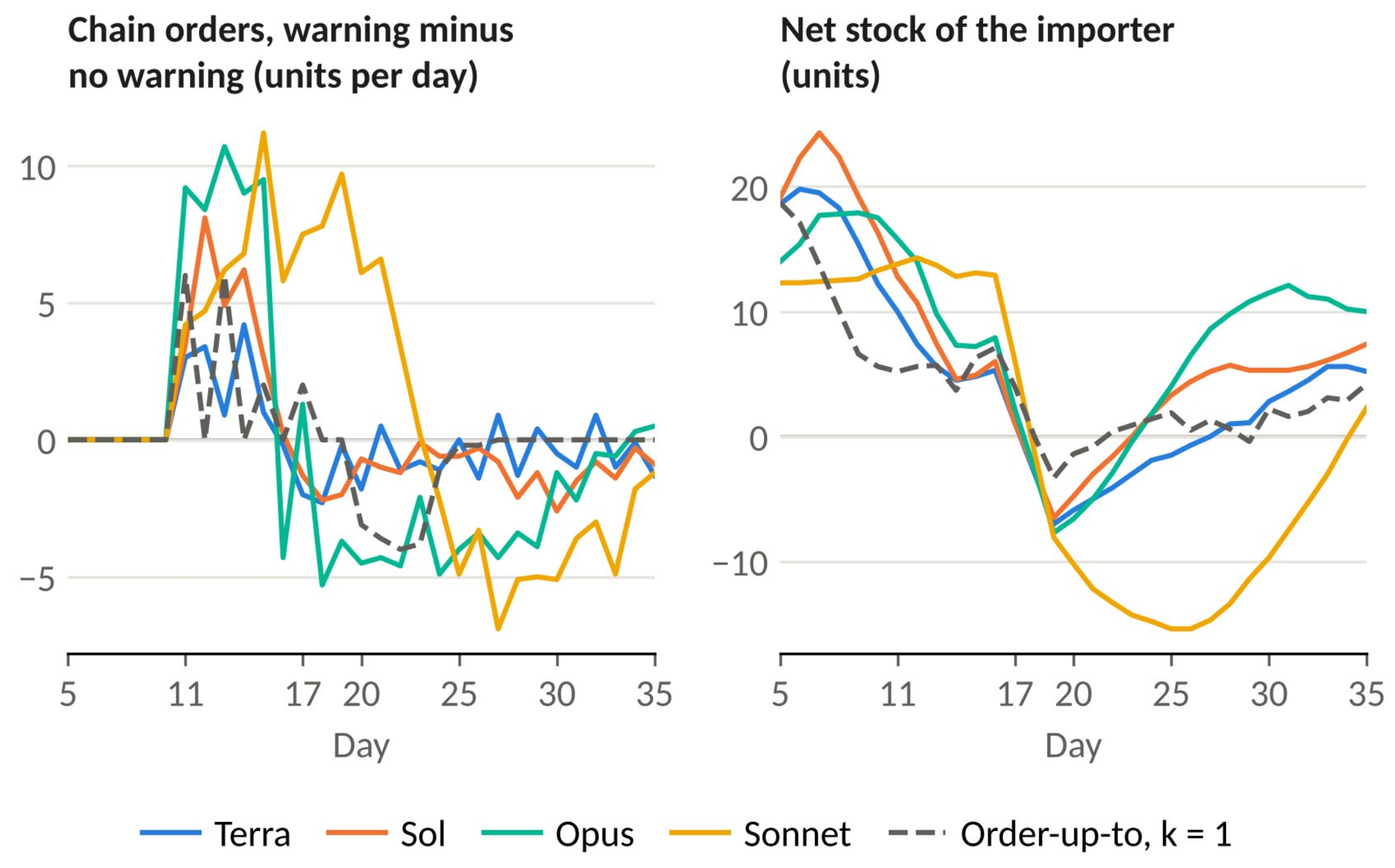

Fig. 2. Extra orders of the whole chain compared with the same suspension without a warning (left) and net stock of the importer (stock on hand minus backlog, right) on the path with a toward update and a suspension, mean of the 10 demand paths. The warning lasts from day 11 to day 15, and the suspension from day 17 to day 19.

## VII. CONCLUSION

### A. Main Findings

This study tested LLM ordering agents in a four-echelon supply chain under a probabilistic typhoon warning, and compared them with benchmark policies that received the same warnings.

RQ1 asked how LLM ordering agents respond to typhoon warnings. Three of the four models over-reacted. The GPT-6.2 Sol, Opus 5.5 and Sonnet 5.5 chains ordered about two to four times as much extra stock as the benchmark that buffers the expected shortage, and even their retailers and wholesalers, which the port does not supply directly, stocked up. The GPT-6.1 Terra chain ordered about as much extra stock as this benchmark. All four chains ordered more on average after an update that the storm was heading for the port than after one that it was turning away, but most of their extra stock had already been ordered under the first warning.

RQ2 asked whether the warning system lowers the expected cost of a supply chain run by LLM agents. There was no evidence of a gain for any of the four LLM chains, although the best benchmark, the order-up-to policy with a buffer of half the expected shortage (k = 0.5) at the importer only, gained a little. The warning raised the cost of the Opus and Sonnet chains on every demand path, and the rise in their expected cost was 2.25 and 6.15 times the typhoon's expected cost to the order-up-to chain without a warning (1.07 and 2.93 times with its expected cost over the 1,000 benchmark paths). The rise was smaller for the Terra and Sol chains, and the Sol chain did significantly worse than the best benchmark ($p = 0.027$). Even when the port suspended, no LLM chain gained on average.

RQ3 asked which ordering behaviors go with a larger change in the cost. The size of the reaction during the warning (days 11 to 15) does not explain the loss of the Sonnet chain, which lost more than order-up-to buffers with the same reaction index (RI) on all 10 demand paths. The loss also went with when the agents ordered, and the chain that lost the most, Sonnet, was the one whose agents counted the least of their supply line, the stock already ordered but not yet received. The Terra, Sol and Opus agents counted 47% to 90% of their supply line. Compared with the Sonnet agents, they placed a larger share of their extra orders on days 11 and 12, the only days of the warning on which an importer's order could still pass the port before it stopped on day 17, and they ordered less during the suspension than without a warning. The Sonnet agents counted at most 27% of their supply line and ordered more during the suspension than without a warning, and on the path with a toward update and a suspension, most of the Sonnet chain's loss came from a shortage at the importer rather than from holding extra stock.

## B. Implications for Research and Practice

First, a warning of a port suspension offers a lean chain, which holds little safety stock, a little to gain and much to lose, so its value depends on how the chain uses it. This extends to port suspensions the principle that the value of a forecast depends on the decision that it feeds [38], [39]. When every echelon stocks up, the retailer and the wholesaler pull stock away from the importer before the suspension. This competition for stock resembles the shortage gaming that Lee et al. [6] traced to rationing, but here a forecast alone sets it off. Lewis et al. [24] showed that the best ordering policy may depend on the state of the port, and our results add that the best buffer before a suspension depends on the forecast and on where the company sits in the chain.

Second, in our four models, the agent that counted least of its supply line, the stock already ordered but not yet received, lost the most from the warning. Earlier studies found that people [13] and LLM agents [10] do not fully count their supply line, and in our results this known bias went with a larger loss from the warning.

Third, three of our four LLM models over-reacted to the warning and none under-reacted, which adds evidence to the over-reaction side of mixed earlier findings. Long et al. [11] added guardrails to keep LLM agents from over-ordering in panic, whereas the models in Kuai et al. [25] often set a lower alert level than the situation required. One reason may be the task. Kuai et al. [25] scored single answers about alert levels and preparedness measures, whereas our agents ordered day after day, so each day under a warning added another extra order.

For practice, a firm that lets LLM agents place its orders can make three checks. The firm should check whether agents that the port does not supply directly stock up anyway, because in the benchmarks only buffers at the importer paid off, and in our chains the retailers and wholesalers stocked up although every prompt showed where the port sits in the chain. Before handing the daily restocking to an agent, the firm can test with an ordering regression like ours whether the agent orders less when its supply line rises. The firm should also test its agents on false alarms, after which the port stays open, and not only on suspensions, because false alarms caused 44% to 90% of the loss in our four LLM chains.

### C. Limitations and Future Research

Several limits point to future research. The supply chain follows the rules and settings of earlier beer-game studies and simulations, with one product, one port, a factory with unlimited capacity and short delays. In our model the port carries only this chain's cargo, so an order above the mean rate after reopening lengthens the chain's own queue. The benchmarks are also simple ordering rules rather than optimal policies. The setting may therefore be far from real supply chains. Testing the same warnings in larger chains with long ocean routes, alternative ports and real terminal records would show how far the results carry over.

The LLM experiment also leaves room for extension. It tested four models from two providers, run at their default sampling and a low reasoning effort, with one prompt wording and 10 demand paths. Ten demand paths separate a large loss from no loss, but not a small gain or loss, which even the best benchmark shows only over 1,000 paths. Further models, prompt wordings, reasoning settings and demand paths would show how widely the results hold. A prompt that shows each agent its inventory position, the stock on hand minus the backlog plus the supply line, would test directly whether counting the supply line drives the loss, and a laboratory experiment with the same warnings would compare people with LLM agents. Even so, to our knowledge this study is the first chain-wide test of LLM ordering agents under a typhoon warning that is updated and may not come true. It also provides benchmarks that show how large a response to the warning should be, and it releases the simulator, the prompts and every model response, so that other ordering agents can be tested on the same benchmark.

## DATA AVAILABILITY

The code, the data, all model responses, the full parameter set, the reference data for the port model and the sensitivity results are available at https://github.com/HanseonJoo/How-LLM-Ordering-Agents-Respond-to-Imperfect-Typhoon-Warnings-in-a-Multi-Echelon-Supply-Chain.

## ACKNOWLEDGMENT

Claude (Anthropic) was used to assist with the simulation and analysis code in the online repository, including the scripts that draw Figs. 1 and 2, with language editing and translation into English throughout the manuscript, and with checking the references.

## APPENDIX A. AGENT PROMPT

Every agent receives the same system prompt. The user prompt below is the importer's prompt on day 16 of the path with a toward update that ends in a suspension. Demand is held at its mean of 4 for readability. Other roles change only the first line and the delay sentence.

```
SYSTEM
You place the daily order for one company in a four-company supply chain.
Reply with one JSON object and no other text: no working, no explanation
outside it.

USER
Role: importer. Supplier: overseas factory. Customer: wholesaler.
Chain: overseas factory → sea, via the import port → importer → wholesaler →
retail store → consumers.
Consumer demand: a random whole number from 0 to 8 each day, all equally
likely. Unfilled orders are backlogged.
Your orders reach the factory in 2 days; its shipments reach you by sea 2
days after it ships, through the import port while the port is open.
Cost per unit per day: 0.5 in stock, 1.0 backlogged. Goal: lowest total cost
of all four companies. The business runs about 60 days.

Day 16. Received 4. Customer order 4. Stock 12. Backlog 0. Ordered, not yet
```

```
received 12. Port: open.
Your record (day: received/customer order/stock/backlog/your order):
0: 4/4/12/0/4
1: 4/4/12/0/4
2: 4/4/12/0/4
3: 4/4/12/0/4
4: 4/4/12/0/4
5: 4/4/12/0/4
6: 4/4/12/0/4
7: 4/4/12/0/4
8: 4/4/12/0/4
9: 4/4/12/0/4
10: 4/4/12/0/4
11: 4/4/12/0/4
12: 4/4/12/0/4
13: 4/4/12/0/4
14: 4/4/12/0/4
15: 4/4/12/0/4
Notices:
[day 11] Typhoon forecast to reach the port area around day 18. Chance the
port suspends for about 3 days, from about a day before landfall: 50%.
[day 15] Update: the track is heading for the port, arrival around day 18.
Chance of the suspension: 70%.
[day 16] Port: full suspension of ships and cargo from tomorrow (day 17),
ahead of landfall around day 18; reopening expected around day 20.
Reply with only this JSON: {"order": <whole number, 0 or more>, "reason":
"<one short sentence>"}
```